\documentclass[times,twocolumn,final,authoryear]{elsarticle}
\usepackage{booktabs}
\usepackage{ycviu}
\usepackage{framed,multirow}
\usepackage[table,xcdraw,dvipsnames]{xcolor}
\usepackage{multirow}
\definecolor{linkcolor}{rgb}{0.3,0.5,0.7}
\usepackage[colorlinks=true, urlcolor=MidnightBlue, citecolor=red]{hyperref}
\AtBeginDocument{\hypersetup{urlcolor=MidnightBlue, citecolor=MidnightBlue, linkcolor=MidnightBlue}} 
\usepackage{amssymb}
\usepackage{latexsym}
\usepackage{soul,color}
\usepackage{url}
\definecolor{newcolor}{rgb}{.8,.349,.1}

\journal{Computer Vision and Image Understanding}

\begin{document}

\clearpage

\ifpreprint
  \setcounter{page}{1}
\else
  \setcounter{page}{1}
\fi

\begin{frontmatter}

\title{Spatio-temporal Features for Generalized Detection of Deepfake Videos}

\author[1]{Ipek \snm{Ganiyusufoglu}\corref{cor1}} 
\cortext[cor1]{Corresponding author: }
\ead{ipekgan@gmail.com}

\author[1,2]{L. Minh \snm{Ng\^{o}}}
\author[1]{Nedko \snm{Savov}}
\author[1]{Sezer \snm{Karaoglu}}
\author[1,2]{Theo \snm{Gevers}}

\address[1]{3DUniversum, Science Park 400, 1098 XH Amsterdam, The Netherlands}
\address[2]{University of Amsterdam, Science Park 904, 1098XH Amsterdam, The Netherlands}

\received{1 May 2013}
\finalform{10 May 2013}
\accepted{13 May 2013}
\availableonline{15 May 2013}
\communicated{S. Sarkar}

\begin{abstract}
For deepfake detection, video-level detectors have not been explored as extensively as image-level detectors, which do not exploit temporal data.


In this paper, we empirically show that existing approaches on image and sequence classifiers generalize poorly to new manipulation techniques. To this end, we propose \textit{spatio-temporal features}, modeled by 3D CNNs, to extend the generalization capabilities to detect new sorts of deepfake videos.

We show that spatial features learn distinct deepfake-method-specific attributes, while spatio-temporal features capture shared attributes between deepfake methods. We provide an in-depth analysis of how the sequential and spatio-temporal video encoders are utilizing temporal information using DFDC dataset \citep{Dolhansky2020TheDataset}. Thus, we unravel that our approach captures local spatio-temporal relations and inconsistencies in the deepfake videos while existing sequence encoders are indifferent to it.

Through large scale experiments conducted on the FF++ \citep{Rossler2019FaceForensics++:Images} and Deeper Forensics \citep{jiang2020deeperforensics10} datasets, we show that our approach outperforms existing methods in terms of generalization capabilities.

\end{abstract}


\end{frontmatter}

\begin{figure*}[]
\centering
\includegraphics[width=\textwidth]{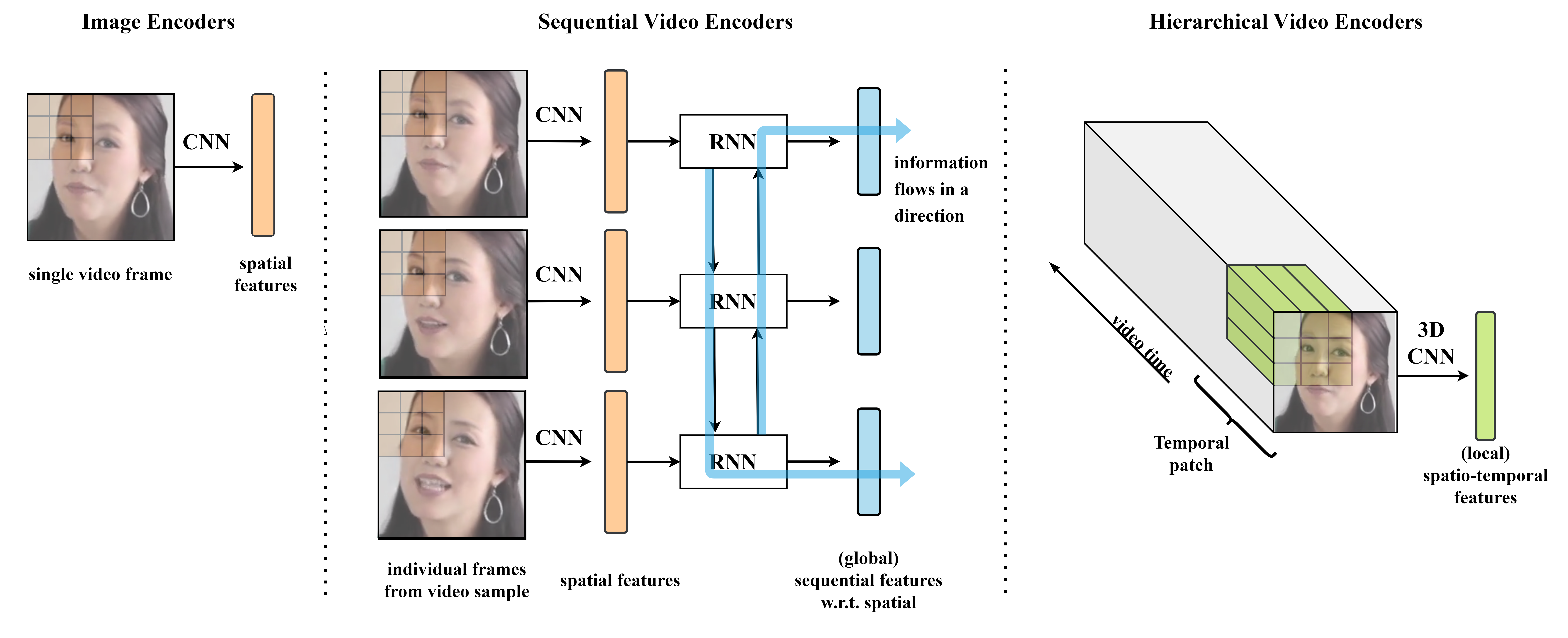}
\caption{Illustration of the differences of image encoders, sequential and hierarchical video encoders. The orange and green squares imply the use of 2D and 3D kernels.}\label{fig:header}
\end{figure*}

\section{Introduction}

Today, an increasing number of media on the Internet is generated by deep learning algorithms. These images often involve manipulation of people's faces, identities or expressions. We commonly refer to such imagery as \emph{deepfakes}.
Deepfake images are becoming more \emph{realistic} and easier to produce using various black-box type algorithms\footnote{github.com/iperov/DeepFaceLab} that do not require any expertise. While this technology can act as a great entertainment tool, it can also be used to spread malicious misinformation\footnote{lionbridge.ai/articles/a-look-at-deepfakes-in-2020/}. 
Therefore, it is important to develop deepfake detection methods to counteract the rapid advancements in the generation of deceptive and misleading deepfakes.


The most prominent deepfake detectors are image-based classifiers \citep{Rossler2019FaceForensics++:Images}, i.e. Convolutional Neural Networks (CNNs), trained with individual portrait images. Image-based detectors rely on cues extracted from a single frame to model spatial features. Typically, these cues appear as color and lighting inconsistencies or generation artifacts.
Perfecting such errors on image-level is far less challenging than producing an entire deepfake video that is temporally coherent. 
To this end, the focus of this paper is on the exploring a solution that fully exploits the broader context of deepfake videos.

To address the lack of utilizing temporal information, \citet{Guera2019DeepfakeNetworks} and \citet{Sabir2019RecurrentVideos} propose sequence classifiers, namely Recurrent Neural Networks (RNNs) for the detection of deepfake videos. However, these models are built on top of image encoders. Hence, the temporal information depends on the quality of the spatial features encoded by the CNN backbone. 

Moreover, a necessity for deepfake detectors is that they should be able to detect new types of deepfakes. Generalization is a vital property in deepfake detection  considering the ever-growing pool of diverse deepfake generation methods such as morphable mask models \citep{Dolhansky2020TheDataset, ThiesFace2Face:Videos} and Autoencoders, GANs, and VAEs \citep{LiFaceShifter:Swapping, NirkinFSGAN:Reenactment, ThiesDeferredTextures, Zakharov2019Few-ShotModels, jiang2020deeperforensics10}. When the detector is trained on a particular set of such manipulation methods, it should be able to transfer its detection capabilities to unseen deepfake algorithms.

In this paper, we empirically show that existing methods on image and sequence classifiers generalize poorly to new manipulation techniques. To this end, we propose \textit{spatio-temporal features}, modeled by 3D CNNs, to extend the generalization capabilities to detect deepfake videos. RNNs and 3D CNNs are both video encoders for temporal modeling. However, they exploit temporal information differently because of their architectural differences. RNNs process a video-clip as a sequence in a particular direction, building up an understanding on top of the preceding time-steps. 3D CNNs view the video input as a hierarchy of temporal patches. They simultaneously learn an integrated representation of the spatial and temporal dimensions instead of learning through two separate components as in sequential video encoders. These differences are illustrated in Figure \ref{fig:header}.
Moreover, because 3D CNNs use 3D kernels convolving over space and time, they are able to capture \textit{local irregularities} within a \textit{few consecutive} frames. In contrast, RNNs capture global properties as they process the sequence at a higher feature level. As a result, 3D CNNs are selected to identify spatio-temporal inconsistencies, cues and patterns to enable generalized deepfake detection.

We empirically compare our approach with two existing approaches, CNNs and RNNs, using well-known datasets in the field \citep{Rossler2019FaceForensics++:Images, Dolhansky2020TheDataset, jiang2020deeperforensics10}.


Our key contributions are as follows:
\begin{itemize}[(a)]
    \item  We propose to exploit spatio-temporal models to enhance the generalization of deepfake detection. 
    \item It is demonstrated that, in contrast to spatial features, spatio-temporal features capture similarities between different types of deepfakes enhancing the generalizing behavior of spatio-temporal models.
    \item Our analysis shows that spatio-temporal video encoders capture local disruptions and inconsistencies in spatio-temporal correlation while sequential video encoders are indifferent to it.
    \item It is empirically validated that our approach outperforms existing methods on detecting unseen deepfake types.
\end{itemize}

\section{Related Work}

\begin{figure}[!t]
\centering
\includegraphics[width=0.45\textwidth]{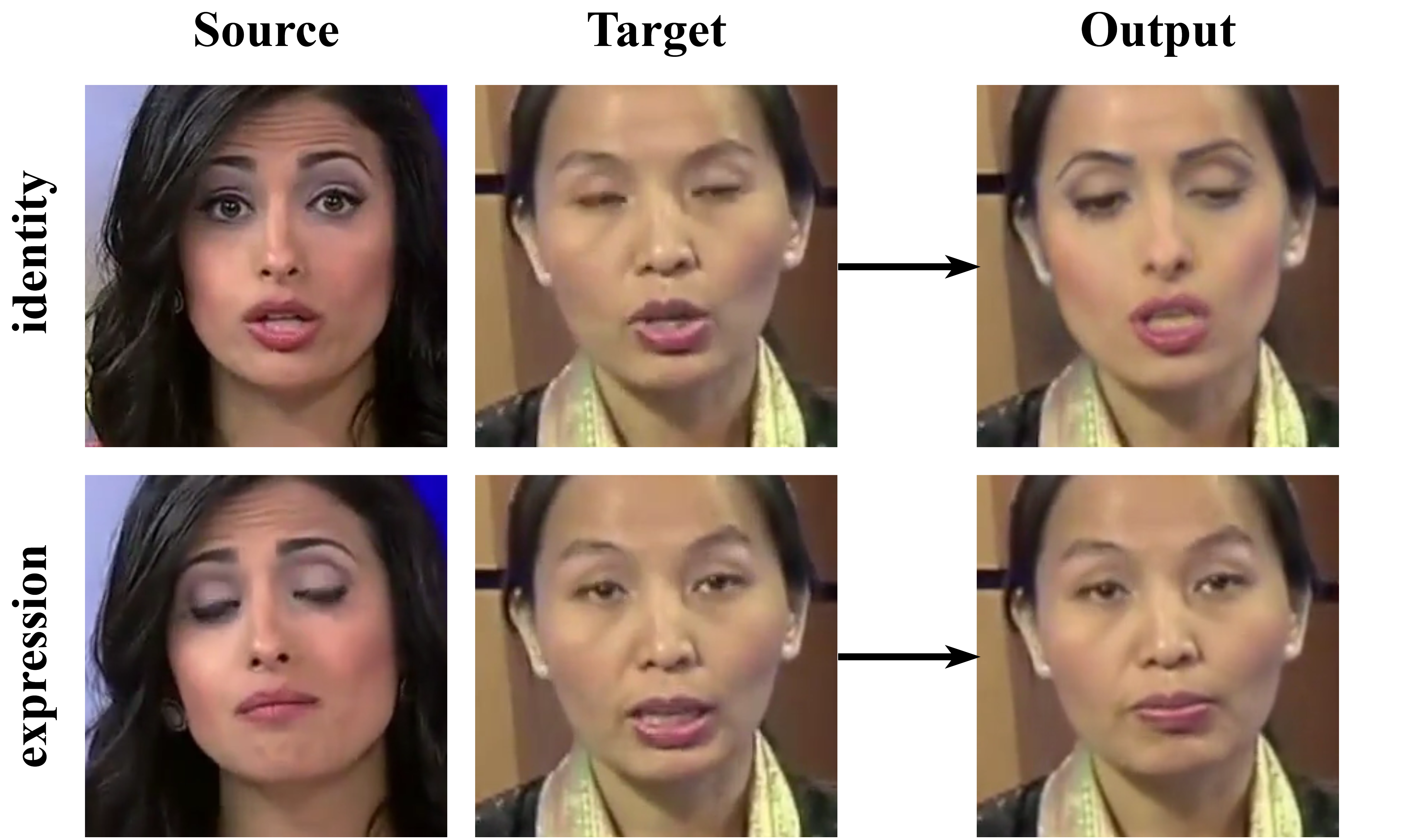}
\caption{Examples of source-to-target transfer methods; identity transfer or face swapping on top and facial expression reenactment at the bottom. The mouth region is reenacted with respect to the source actor. Images are taken from the FaceForensics++ dataset \citep{Rossler2019FaceForensics++:Images}.} 
\label{fig:explain}
\end{figure}

\paragraph{Deepfake types and generation} 
In this paper, we focus on deepfakes generated by a source-to-target transfer i.e. facial swaps and reenactment. This is because these methods manipulate imagery of existing people rather than generating fake people that do not exist\footnote{thispersondoesnotexist.com}. Therefore, source-to-target transfer methods are more suitable for misinformation attacks against public figures. Facial swapping refers to when the likeness in a target video is replaced by a source face. Facial reenactment refers to when the target actor's expression and or pose is manipulated to follow the source actor. Examples of source-to-target transfer methods are shown in Figure \ref{fig:explain}. Both kinds of deepfakes can be generated by a large variety of methods such as morphable mask models \citep{ThiesFace2Face:Videos} or generative modeling approaches including Autoencoders, VAEs, and GANs, \citep{LiFaceShifter:Swapping, NirkinFSGAN:Reenactment, ThiesDeferredTextures, Zakharov2019Few-ShotModels, jiang2020deeperforensics10}.

\paragraph{Deepfake Detection Benchmarks}
Today, the most popular deepfake detection benchmark is FaceForensics++ (FF++) \citep{Rossler2019FaceForensics++:Images}, a small dataset generated from 1000 real videos and 5 deepfake generation methods; Face2Face \citep{ThiesFace2Face:Videos}, Deepfakes\footnote{https://github.com/deepfakes/faceswap}, FaceSwap\footnote{https://github.com/MarekKowalski/FaceSwap/}, NeuralTextures \citep{ThiesDeferredTextures}, FaceShifter \citep{LiFaceShifter:Swapping} applied to each video. There are other more recently released benchmarks as well, namely, Deepfake Detection Challenge (DFDC) \citep{Dolhansky2020TheDataset} and Deeper Forensics \citep{jiang2020deeperforensics10}. However, they are, as of yet, not as prominent as FF++. The former dataset, DFDC, is a much larger dataset of approximately 100K fake videos that contain more diversity in terms of identity, expression, pose, motion, environment, and illumination. The latter dataset expands the FF++ benchmark to include a new generation method and real-world perturbations applied to those images. 

\paragraph{Image-based Detectors}
There are different image classifier methods to detect deepfakes. However, a number of them rely on heuristically tailored input and therefore are easily to be tricked or obsolete, \citep[e.g.][]{Yang2018ExposingPoses, Zhang2019DetectingVersion}. \citet{Rossler2019FaceForensics++:Images} implement four portrait-image-based detection models; CNNs, and evaluate on the FF++ benchmark dataset. They conclude that XceptionNet \citep{Chollet2016Xception:Convolutions} architecture is the best performing detector across all four methods and compression perturbations, thus established the state-of-the-art at the time. More recently, for the FF++ benchmark, EfficientNets \citep{TanEfficientNet:Networks} show to be promising detectors. Although these image-based classifiers have proven to be useful, by design, they are limited to gather intra-frame cues and inconsistencies and are not able to capture spatio-temporal relations. Therefore, these models are limited in exploiting the temporal information to detect deepfake videos. 
 
\paragraph{Sequence-based Detectors}
Detection of video sequences has become possible with the release of the FaceForensics benchmarks \citep{Rossler2019FaceForensics++:Images}. \citet{Sabir2019RecurrentVideos} use recurrent convolutional models, where video frames are first passed through a CNN to extract spatial features, and a RNN then models the sequence of these features. They use DenseNet \citep{Huang2016DenselyNetworks} and a bi-directional GRU \citep{cho2014learning} respectively. \citet{Guera2019DeepfakeNetworks} explores another RNN approach. The pipeline is very similar but they use Inception-v3 \citep{Szegedy2016RethinkingVision} for the convolutional component and LSTM \citep{HochSchm97} as the recurrent unit.

\paragraph{Video Encoding}
Video encoders can be clustered in two families; sequential and hierarchical video encoders, \citep{Ghodrati2018VideoEvaluation}. The former uses RNN-based encoders, while the latter entails 3D CNNs. 

RNNs apply the same function iteratively to the time-steps of an input sequence. They preserve an internal memory of the past steps; thus, a representation of the preceding sequence is kept and combined with the next step. This way, RNNs model the transition functions from preceding states to the current state. 

In 3D CNNs, 3D kernels are convolved over consecutive frames of video clips in dimensions width, height and time. The 3D convolutional temporal information processing is significantly different than RNNs. Firstly, RNN-based video encoders process videos sequentially in a direction while accumulating a global state representation, while 3D CNNs view the video as a hierarchy of temporal patches, thus capturing local spatio-temporal features. Secondly, RNN-based models first learn spatial features then model sequential relations between them, thus they learn the video encodings in two steps. In contrast, 3D CNNs convolve over both space and time dimensions simultaneously through-out the architecture, thus learn a joint representation of them. 

The advantage of using either of these architectures depends on the context and the data of the task at hand. This is because certain temporal processing traits of the architecture can be more beneficial with respect to a task. For example, \citet{Ghodrati2018VideoEvaluation} and \citet{ManttariINTERPRETINGNETWORKS} provides comparisons of RNN and 3D CNN architectures in the context of action recognition. In their analysis, \citet{Ghodrati2018VideoEvaluation} evaluate the two video encoders for three different subtasks that test the encoders' ability to capture temporal asymmetry, continuity or causality of a video. They found that in tasks that require an understanding of direction and asymmetry; predicting if a video is playing forward or backward, RNN-based models outperform 3D CNNs. This is inline with RNNs that process videos sequentially in a direction. However, in other tasks, where the entire sequence direction does not carry any information, they found that 3D CNN outperforms the RNN-based models. Deepfake videos are ultimately recordings of people's faces. Hence the data is quite monotone in contrast to action data and the direction or the events occurring in the video are not semantically relevant to classify whether the video is real or manipulated. The deepfake cues are rather hidden in spatio-temporal correlations of a few consecutive frames. Therefore, based on the described nature of deepfake video data and \citet{Ghodrati2018VideoEvaluation}'s findings, we find 3D CNNs to be more suitable for this task.

 

\section{Method}
We propose spatio-temporal learning for the generalized detection of deepfake videos. We expect spatio-temporal modeling to capture more local cues and patterns in deepfake videos allowing them to identify more types of deepfakes and thus be more generalizing. We evaluate our approach in contrast to existing baselines that represent spatial and sequential learning.

\subsection{Spatio-temporal Models}
Spatio-temporal modeling is obtained by the use of 3D CNNs. The model learns a joint representation of the spatial and temporal features throughout the architecture. 

\paragraph{Model Choice} To represent spatio-temporal learning, we use two different 3D CNNs, originally used for action recognition. The primary model is R3D-18 from the work of \citet{Tran2017ARecognition} where they demonstrate that spatio-temporal video encoding is valuable by showing that 3D convolutional residual networks outperform their original 2D architecture in the context of action recognition. R3D-18 is the 3D convolutional version of the 18 layer ResNet \citep{He2016DeepRecognition}. As an auxiliary model, we use I3D-RGB \citep{CarreiraQuoDataset} to demonstrate that our insights hold for different 3D CNN models. I3D consists of two streams of a 3D convolutional version of the Inception-v1, where one stream learns RGB frames and the other learns optical flow. We use only a single stream for RGB to hold all models to the same conditions.

\begin{figure}
\centering
\includegraphics[width=0.48\textwidth]{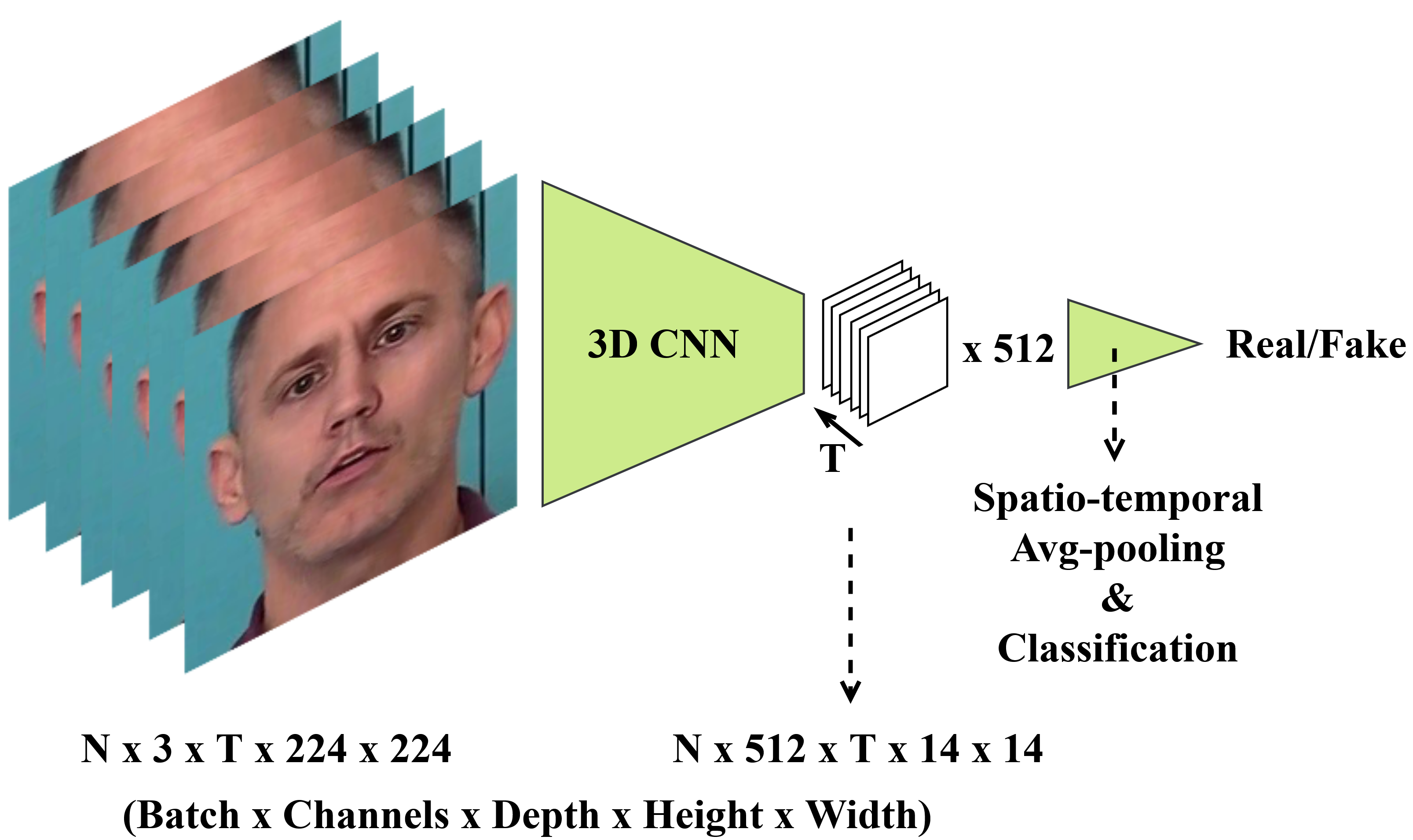}
\caption{Example input and highest feature map dimensions for a 3D CNN pipeline (based on R3D \citep{Tran2017ARecognition}) without temporal downsampling.} 
\label{fig:3dnet}
\end{figure}

\paragraph{Removing Temporal Downsampling}
Regular 2D CNNs have spatial downsampling. This makes training faster and memory requirements lighter while allowing the network to ignore redundant information. R3D uses convolutional strides for spatio-temporal downsampling. This occurs at three stages in the network; each time, the temporal dimension is reduced to half its size. Especially at higher levels, the remaining temporal dimension is barely kept depending on input depth. We want to allow all network levels to learn from the temporal data without compressing it into a smaller representation. With this motivation, we remove temporal downsampling by using a stride of 1 for depth. 
This leaves us with the example 3D pipeline visualized in Figure \ref{fig:3dnet} with example input dimensions and resulting output feature maps. 

\subsection{Baselines}
\paragraph{Spatial modeling} As a primary baseline and representation for learning purely spatial features with no temporal awareness, we use two image encoders. The first one is  XceptionNet \citep{Rossler2019FaceForensics++:Images} representing the previous state-of-the-art for Deepfake detection. The second image encoder baseline is EfficientNet-B3 \citep{TanEfficientNet:Networks} that slightly outperforms Xception on the ImageNet benchmark \footnote{https://paperswithcode.com/sota/image-classification-on-imagenet}. Having two different state-of-the-art networks allows us to verify that certain behaviors are more characteristic to CNNs rather than specific to a single network.

After the image classifiers are trained with individual frames of the videos, they are used frame-by-frame on videos to acquire a group of probability predictions. These predictions are averaged to get a single final score for a video sample.

\paragraph{Sequential modeling} 

As a secondary baseline, we use two sequence encoder pipelines  \citep{Guera2019DeepfakeNetworks, Sabir2019RecurrentVideos}. For these, we use the image baseline EfficientNet-B3 as the backbone feature extractor, followed by an RNN. The last timestep of the sequential encoding acquired from the RNN is taken and forwarded through a fully connected network to get a single prediction.

The first pipeline uses the LSTM similar to \citet{Guera2019DeepfakeNetworks}. We have a two-layer mono-directional LSTM followed by a three layer fully-connected network with ReLU non-linearity. There are two Dropout layers, first after the LSTM and the second before the second linear layer. We use EfficientNet image baseline's trained weights to initialize the backbone and freeze it completely. 

The second pipeline is almost identical to \citet{Sabir2019RecurrentVideos}. We use a single-layer bi-directional GRU, followed by a Dropout layer and a single output layer. We backward the loss on only the output of the final timestep. In this pipeline, after initializing EfficientNet-B3 with newly pretrained weights, we freeze the first 20 out of 25 of its convolutional blocks.

\section{Experiments}
This section presents multiple experiments to analyze the differences of spatial, sequential, and spatio-temporal learning, including their generalization to new deepfake methods. 

\begin{table*}[!t]
\caption{Test precision per fake classes when all methods are included in training (left side) and when a single method is left out of training and validation but included in testing (right side). for FF++ test set. Different colors indicate that the values are from different runs, separately for each model. Final column on bottom table shows the drop in average accuracy. (DF: Deepfakes, F2F: Face2Face, FS: FaceSwap, NT: NeuralTextures, FShi: FaceShifter). The RNN model here is bi-GRU. A full version of the table can be found in supplementary material \ref{supp}.}
\label{genel}
\centering
\footnotesize
\begin{tabular}{c|l|ccccc||c|c|cccccc}
\hline
\multicolumn{1}{c|}{} &  & \multicolumn{6}{c|}{all methods included} & \multicolumn{1}{c|}{} & \multicolumn{6}{c}{one method is left-out} \\ \cline{3-15} 
\multicolumn{1}{l|}{\multirow{-2}{*}{encoder type}} & \multirow{-2}{*}{model} & DF & F2F & FS & NT & FShi & overall avg & \multicolumn{1}{l|}{} & DF & F2F & FS & NT & \multicolumn{1}{c||}{FShi} & drop \\ \hline
 & Xcept. & 99.29 & 97.86 & 97.86 & 95.71 & 95.71 & 98.64 &  & \cellcolor[HTML]{E7FFFE}21.43 & \cellcolor[HTML]{FFEEED}12.14 & \cellcolor[HTML]{FFEDD7}0.00 & \cellcolor[HTML]{E0FFE0}0.00 & \multicolumn{1}{c||}{\cellcolor[HTML]{E9EBFF}1.43} & -10.28 \\
\multirow{-2}{*}{image-based} & Eff-B3 & 97.86 & 94.29 & 98.57 & 97.86 & 96.43 & 98.50 &  & \cellcolor[HTML]{E7FFFE}15.71 & \cellcolor[HTML]{FFEEED}7.14 & \cellcolor[HTML]{FFEDD7}0.00 & \cellcolor[HTML]{E0FFE0}0.00 & \multicolumn{1}{c||}{\cellcolor[HTML]{E9EBFF}15.00} & -10.49 \\ \hline
sequential & RNN & 100 & 100 & 100 & 98.57 & 98.57 & 98.29 &  & \cellcolor[HTML]{E7FFFE}71.43 & \cellcolor[HTML]{FFEEED}35.00 & \cellcolor[HTML]{FFEDD7}4.29 & \cellcolor[HTML]{E0FFE0}7.86 & \multicolumn{1}{c||}{\cellcolor[HTML]{E9EBFF}26.43} & -6.99 \\ \hline
spatio-temporal & R3D & 97.86 & 97.86 & 99.29 & 99.29 & 95.71 & 95.79 &  & \cellcolor[HTML]{E7FFFE}62.14 & \cellcolor[HTML]{FFEEED}75.00 & \cellcolor[HTML]{FFEDD7}5.00 & \cellcolor[HTML]{E0FFE0}57.14 & \multicolumn{1}{c||}{\cellcolor[HTML]{E9EBFF}56.43} & -4.90 \\
(proposed) & I3D & 97.86 & 98.57 & 94.29 & 95.00 & 96.43 & 88.57 &  & \cellcolor[HTML]{E7FFFE}77.86 & \cellcolor[HTML]{FFEEED}87.86 & \cellcolor[HTML]{FFEDD7}12.86 & \cellcolor[HTML]{E0FFE0}71.43 & \multicolumn{1}{c||}{\cellcolor[HTML]{E9EBFF}63.57} & -3.37 \\ \hline
\end{tabular}
\end{table*}

\subsection{Setup}

\paragraph{Data}
In this paper, we primarily use the FF++ \citep{Rossler2019FaceForensics++:Images} dataset. Furthermore, we use the DFDC \citep{Dolhansky2020TheDataset} and Deeper Forensics \citep{jiang2020deeperforensics10} datasets. All three datasets consist of videos. For FF++, we use the lightly compressed version (c23) and split the original 1000 real videos into train, validation and test sets of respectively 720, 140 and 140 videos similar to \citet{Rossler2019FaceForensics++:Images}. The DFDC dataset is provided in folders where each folder consists of a number of actors unique to that folder. At the time of this research, the test set was not public, but has been published recently. Therefore, we split the main dataset by these folders and sampled around 50.000 fake and 10.000 real samples for training from 40 arbitrary folders, and sampled one and five thousand videos for validation and testing from 5 different folders each. We do not use Deeper Forensics for training, we only test on the 00 and end-to-end labeled portion of it with no perturbations, using the same split as FF++. 

\paragraph{Data processing} We use face detection \citep{JMLR:v10:king09a} and landmark alignment \citep{bulat2017far} to extract square facial crops of faces, with an additional margin of 40 percent of the detected bounding box size. We extract faces from 3 frames per second. This can result in a different number of frames as the video samples can have different duration. Face detection can result in false faces or scales. To correct such errors, we apply two filtering steps using the intersection over union of boxes and size outlier detection. We detail these steps in the supplementary material \ref{supp}.


\paragraph{Input}
The video encoders take 16 frame sequences as input, specifically a 4D tensor of dimensions $3\times 16 \times 224\times 224$. Sixteen frames roughly correspond to 5 seconds of a video. At training time, a random 16-frame consecutive subsequence is sampled from the available images for a video sample. If the image count is less than 16, the video is looped until it suffices. The sequence encoders forward the video-tensor images independently through the CNN component and then stack the features back into a sequence for the RNN part. The 3D CNNs take the 4D input as is. The image encoders are trained with batches of frames from different videos. All models except for XceptionNet use 224 for image dimensions and ImageNet values for mean and standard deviation normalization of the data. XceptionNet uses 299 and 0.5 for all normalization values.

\paragraph{Training} During training, we use balanced batches with an equal amount of real and fake samples. For FF++, we try to increase the diversity of a batch by also iterating through a shuffled order of different deepfake methods so that certain methods do not dominate batches. For DFDC, as deepfake labels are unknown, we iterate through the folders to provide a diverse set of identities. All models are ultimately binary classifiers; they optimize binary cross-entropy (log-loss) and predict the probability of being fake.

 \paragraph{Inference} When testing video encoders, we go over all available video frames in a sliding window manner and average predictions. For image encoders, we average the per-frame predictions for the entire clip.

\paragraph{Implementation}
We use PyTorch implementations for all models, where pre-trained weights are also provided. Only I3D weights are separately acquired\footnote{https://github.com/piergiaj/pytorch-i3d} for the Charades\footnote{https://prior.allenai.org/projects/charades} dataset for action recognition. For R3D, we use weights pre-trained on the action recognition dataset Kinetics-400 \citep{Kay2017TheDataset}.

\paragraph{Evaluation}
We follow the practice of \citet{Rossler2019FaceForensics++:Images} and use accuracy for evaluation of the detectors. The accuracy is observed at class-level, i.e. real class precision or precision of each of the different fake methods. The overall accuracy is calculated as an average of real and fake-class accuracies to simulate a balanced test set.

\paragraph{Reproducibility}
We provide further details on the training and testing processes, input clips, data augmentations, optimizers and parameter settings in the supplementary material \ref{supp}.

\subsection{Generalization to Unseen Deepfake Algorithms} \label{exgenel}
Generalization is an essential quality of a deepfake detector. In this experiment, we explore if the detectors can generalize, i.e., whether models can detect the output of new  deepfake generation methods that were not included in training. We use the FF++ dataset as it is labeled by the corresponding fake methods. We train models by leaving one generation method out of training and validation. Then we evaluate on the left-out deepfake method. If a model can generalize, it should show higher accuracy on the excluded samples than a model that cannot.

In Table \ref{genel}, we can see the results of the generalization study for the existing image and sequence encoder baselines and the two 3D video encoders as our approach. The left-side shows the precision per fake-method when the entire training data with \textit{all five} generation methods is used together. The right-side with colored columns shows the results when a single deepfake method is left-out. Each column of a model row corresponds to a different run where that column's method is left out, as implied by the colors. 

Firstly we see that the image encoders, XceptionNet and EfficientNet, have worse than random scores on all left-out methods. This means that most fake images remain undetected. The image encoders confuse the unseen fake types with the real class. Thus if these detectors were to be deployed,  they would be tricked by an unseen deepfake algorithm.  The image encoders fail to detect even the left-out methods that share some traits with the seen methods,  such as architectural or task-based similarities. Most of the misclassifications are also confident probability guesses instead of unsure and close to random.  This means that the models strongly think that unseen methods are not fake.

Secondly, we look at the sequence encoder, RNN. Overall, the sequence encoder classifies more samples correctly compared to the image encoders. However, the detection precision is mostly still worse than random. The sequence encoder being slightly more generalizable shows us that there is additional value in modeling temporal relations.

Thirdly, we examine our proposed method; spatio-temporal modeling as represented by the 3D Convnets R3D and I3D. R3D achieves precision better than random for all manipulation methods except for FS. Besides the FS samples, R3D can generalize to unseen fake types consistently. This implies that spatio-temporal models are able to transfer some of their understanding of already learned fake methods on to new unseen fake kinds, while spatial and sequential models cannot. This observation also holds for I3D, validating that the property is of the spatio-temporal and 3D convolutional learning and not of any other attribute of the R3D network.

It is worth noting that all models ultimately fail to recognize the FaceSwap (FS) samples when it is excluded from training. This shows that FaceSwap is unique and dissimilar to  other methods in the dataset and, therefore, valuable for training. This motivates future training sets in deepfake detection to include a wide variety of deepfakes.

So far, in this experiment, we have discussed the fake class loss because we focused on the evaluation
of left-out methods. We also define a single metric to account for both real and fake class
performance equally and summarize the results in Table \ref{genel}. We calculate how much the overall average accuracy, seen on the top table's rightmost column, dropped for each run where we leave a method out. We average the drop of each run for the left-out methods and report this number on the rightmost column at the bottom table. This metric supports our observations for all of the models. Image encoders show the highest drop, sequence encoders show slightly less drop, while 3D Convnets show the least average drop in accuracy. All available evaluation results of all classes are given in supplementary material.

\subsubsection{Cross-dataset Generalization}

We also test the FF++ trained models on another dataset; Deeper Forensics. This dataset uses the same original 1000 real videos in FF++ but a new deepfake generation method based on variational autoencoders (VAEs) for face swaps. The results, as shown in Table \ref{dfx}, are consistent on this data as well; spatio-temporal models are better at generalizing compared to the existing solutions in question.

\begin{table}[]
\caption{Evaluation of FF++-trained models on the VAE-based method introduced in the Deeper Forensics dataset. The instances of the models tested here are from table \ref{genel}, leftside.}\label{dfx}
\centering
\footnotesize
\begin{tabular}{l|c}
\hline
models & Deeper Forensics \\ \hline
Xcept & 24.28 \\ 
Eff-B3 & 16.43 \\ 
RNN & 18.57 \\ 
R3D & 56.43 \\ 
I3D & 78.57 \\ \hline
\end{tabular}
\end{table}

\subsubsection{Generalization to New Tasks and Architectures}
We provide additional experiments to explore more challenging generalization conditions. We investigate if the models can transfer the knowledge they learn when trained on a specific task; reenactment or swapping, to the other. Similarly, when they are only trained on the generation architectures that synthesize their output using decoders, can the models still detect the output of morphable-mask models that do not utilize any convolutional network for synthesis?

We group the methods based on the task and architecture they involve, as shown by colors in Table \ref{genel2}. DF, NT, and Fshi all involve decoder networks for synthesis, while F2F and FS are morphable mask models and do not perform synthesis. DF, FS and Fshi perform facial swapping while NT and F2F perform reenactment. We leave-out methods from training as a group.

Firstly, we should note that models naturally fail the FS method, as we have discussed previously. Besides that, we see that R3D is performing worse than the previous generalization experiment as now there is less information shared between the seen and unseen deepfake methods. However, R3D still outperforms the image and sequence encoders on average.

Interestingly, we see that in some fake methods, the image and sequence encoder benefited from the exclusion of more methods. For instance, EfficientNet recognizes more F2F samples than before, when FS is also left out. This is also observable for the DF method, which is detected more by image and sequence encoders if other swapping methods are unseen as well, as seen in the bottom table. This may be because the inclusion of different fakes affect the features that the model learns. FS samples are less realistic than DF. Thus with the context of FS in learning, DF may appear on the more realistic side. Alternatively, when a method that shares strong similarities with another method is left-out, this can lead to the model finding new associations between other methods. Thus inclusion of methods will have certain effects on learning, which for instance for the image encoder seems to have a negative effect. However, this is not the case with the 3D video encoders, as they again achieve consistently better generalization performance than the image and sequence encoders.

\begin{table}[]
\caption{Precision per fake-class is shown as in table \ref{genel}. The methods that have the same color were left-out together. On the top, architecture-based and on the bottom task-based generalization are evaluated. (MM: F2F \& FS; learned: DF, NT, FShi; swap: DF, FS, FShi; reenact: F2F, NT)}\label{genel2}
\centering
\footnotesize
\begin{tabular}{l|c|c|c|c|c||cc} 
\hline
\multicolumn{1}{c|}{\multirow{2}{*}{models}} & \multicolumn{5}{c||}{fake generation methods}                                                                                                                                                         & \multicolumn{2}{c}{left-out avg}             \\ 
\cline{2-8}
\multicolumn{1}{c|}{}                        & DF                                    & F2F                                   & FS                                    & NT                                    & FShi                                  & MM                   & learned               \\ 
\hline
Xcept                                        & {\cellcolor[rgb]{0.906,1,0.996}}2.14  & {\cellcolor[rgb]{1,0.933,0.929}}28.57 & {\cellcolor[rgb]{1,0.933,0.929}}0.71  & {\cellcolor[rgb]{0.906,1,0.996}}0     & {\cellcolor[rgb]{0.906,1,0.996}}0.71  & 14.64                & 0.95                  \\
Eff-B3                                       & {\cellcolor[rgb]{0.906,1,0.996}}2.14  & {\cellcolor[rgb]{1,0.933,0.929}}32.14 & {\cellcolor[rgb]{1,0.933,0.929}}2.86  & {\cellcolor[rgb]{0.906,1,0.996}}0     & {\cellcolor[rgb]{0.906,1,0.996}}0     & 17.50                & 0.71                  \\
\hline
RNN                                          & {\cellcolor[rgb]{0.906,1,0.996}}10.00 & {\cellcolor[rgb]{1,0.933,0.929}}26.43 & {\cellcolor[rgb]{1,0.933,0.929}}2.14  & {\cellcolor[rgb]{0.906,1,0.996}}2.14  & {\cellcolor[rgb]{0.906,1,0.996}}0     & 14.29                & 4.05                  \\
\hline
R3D                                          & {\cellcolor[rgb]{0.906,1,0.996}}2.14  & {\cellcolor[rgb]{1,0.933,0.929}}75.71 & {\cellcolor[rgb]{1,0.933,0.929}}0.71  & {\cellcolor[rgb]{0.906,1,0.996}}17.86 & {\cellcolor[rgb]{0.906,1,0.996}}0     & 38.21                & 6.67                  \\
I3D                                          & {\cellcolor[rgb]{0.906,1,0.996}}14.29 & {\cellcolor[rgb]{1,0.933,0.929}}85.71 & {\cellcolor[rgb]{1,0.933,0.929}}18.57 & {\cellcolor[rgb]{0.906,1,0.996}}36.43 & {\cellcolor[rgb]{0.906,1,0.996}}17.14 & 52.14                & 22.62                 \\ 
\hline
\multicolumn{1}{l}{}                         & \multicolumn{1}{l}{}                  & \multicolumn{1}{c}{}                  & \multicolumn{1}{c}{}                  & \multicolumn{1}{l}{}                  & \multicolumn{1}{l}{}                  & \multicolumn{1}{l}{} & \multicolumn{1}{l}{}  \\ 
\hline
\multicolumn{1}{c|}{\multirow{2}{*}{models}} & \multicolumn{5}{c||}{fake generation methods}                                                                                                                                                         & \multicolumn{2}{c}{left-out avg}             \\ 
\cline{2-8}
\multicolumn{1}{c|}{}                        & DF                                    & F2F                                   & FS                                    & NT                                    & FShi                                  & swap                 & reenact               \\ 
\hline
Xcept                                        & {\cellcolor[rgb]{0.906,1,0.996}}30.71 & {\cellcolor[rgb]{1,0.933,0.929}}7.14  & {\cellcolor[rgb]{0.906,1,0.996}}0.71  & {\cellcolor[rgb]{1,0.933,0.929}}6.43  & {\cellcolor[rgb]{0.906,1,0.996}}4.29  & 11.90                & 6.79                  \\
Eff-B3                                       & {\cellcolor[rgb]{0.906,1,0.996}}40.71 & {\cellcolor[rgb]{1,0.933,0.929}}6.43  & {\cellcolor[rgb]{0.906,1,0.996}}0     & {\cellcolor[rgb]{1,0.933,0.929}}5     & {\cellcolor[rgb]{0.906,1,0.996}}1.43  & 14.05                & 5.71                  \\
\hline
RNN                                          & {\cellcolor[rgb]{0.906,1,0.996}}41.43 & {\cellcolor[rgb]{1,0.933,0.929}}9.29  & {\cellcolor[rgb]{0.906,1,0.996}}0     & {\cellcolor[rgb]{1,0.933,0.929}}3.57  & {\cellcolor[rgb]{0.906,1,0.996}}5.71  & 15.71                & 6.43                  \\
\hline
R3D                                          & {\cellcolor[rgb]{0.906,1,0.996}}42.14 & {\cellcolor[rgb]{1,0.933,0.929}}36.43 & {\cellcolor[rgb]{0.906,1,0.996}}5     & {\cellcolor[rgb]{1,0.933,0.929}}40.71 & {\cellcolor[rgb]{0.906,1,0.996}}16.43 & 21.19                & 38.57                 \\
I3D                                          & {\cellcolor[rgb]{0.906,1,0.996}}67.14 & {\cellcolor[rgb]{1,0.933,0.929}}33.57 & {\cellcolor[rgb]{0.906,1,0.996}}12.86 & {\cellcolor[rgb]{1,0.933,0.929}}32.86 & {\cellcolor[rgb]{0.906,1,0.996}}42.86 & 40.95                & 33.21                 \\
\hline
\end{tabular}
\end{table}



To conclude the generalization experiment, we demonstrated that image encoders, i.e., spatial features are not suitable for learning generalizing features across different manipulation methods. Sequential encodings of these features are already a step in the right direction. In comparison, spatio-temporal features of 3D Convnets appears as the most generalizing features and more robust against unseen manipulation kinds.

\begin{figure*}[]
\centering
\includegraphics[width=\textwidth]{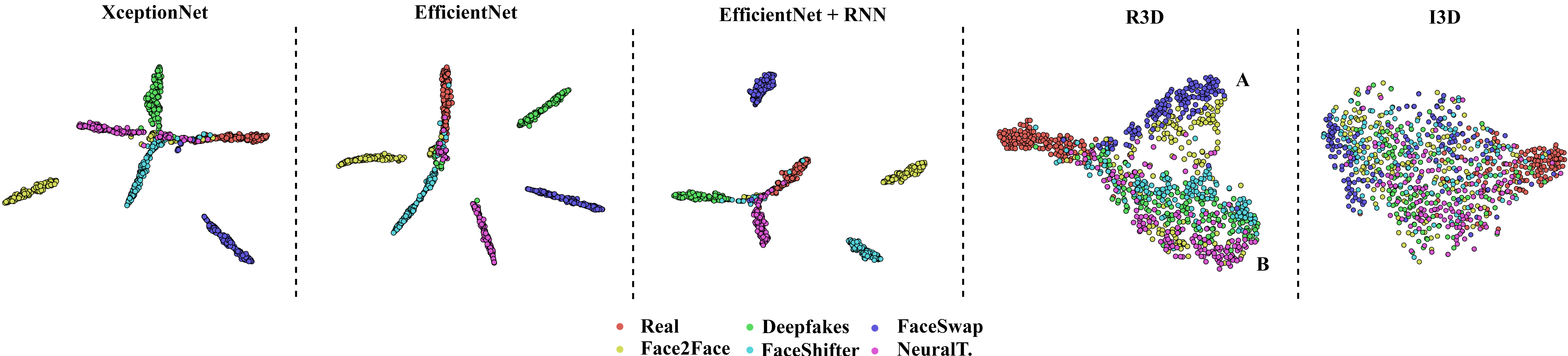}
\caption{t-SNE visualizations of features extracted from models before classification on FF++ test set. t-SNE is computed with perplexity 40, showing results at 2500 iterations. A,B clusters of R3D are marked for referring to in analysis. The figure aims to reveal the different feature relations and similarities between fake methods the models have learned. Three models to the left appear strongly differentiate the fake methods, while two models to the right appear to associate and group fake methods.}\label{tsne}
\end{figure*}

\subsection{Feature Visualization}\label{exfeat}
Image, sequence, and 3D video encoders respectively model spatial, sequential, and spatio-temporal features. In this experiment, we
investigate whether the three types of features capture different information. To achieve this, we examine how the learned feature encodings are distributed using t-SNE \citep{VanDerMaaten2008VisualizingT-SNE}. We acquire feature vectors of FF++ test samples, by extracting the input to the classification layer of each network to have the most descriptive representations.

For this experiment, only the first 16 frames of videos are used to acquire features. For image classifiers, descriptors of all frames are averaged, and for video encoders, only one vector is acquired for the 16-frame input clip. It is also important to emphasize that models are supervised only by real and fake labels, not by the specific fake methods.

As t-SNE is a stochastic algorithm with parameters, it should be examined at different perplexity values, random seeds, and iterations. We test a range of values and different seeds
and observe that the clustering behavior is entirely consistent. The t-SNE results for the image-, sequence- and video-encoders and the final parameter settings can be found under Figure \ref{tsne}. The data samples are colored by real class and specific fake methods. 

We see that the image encoders, XceptionNet and EfficientNet, can strongly differentiate each manipulation type. Considering the models are not supervised by fine-grained fake labels, this implies that the image encoders capture unique cues and patterns specific to types of deepfakes. 

Sequence descriptors of the RNN model also result in strong and distinct clusters similar to the image encodings, which is expected as they build the sequential understanding from the image encodings of EfficientNet backbone.

For R3D, there are no six distinct clusters. Instead, we see that multiple fake classes appear in large clusters, marked with A and B. In cluster A, Face2Face and FaceSwap appear closely grouped. Both of these methods involve morphable masks, and no
synthesis. In cluster B, we mainly see three classes; FaceShifter, Deepfakes, and NeuralTextures, all of which synthesize their output, i.e., involve decoder networks. In this cluster, there is also a notable amount of Face2Face-samples, which appear mostly together with NeuralTextures. This is because both are reenactment methods and can share some features resulting from their task.

We see that the secondary 3D Convnet, I3D, exhibits clustering behavior similar to R3D. The fake samples mostly appear in a large cluster, in which some methods remain more concentrated, like FaceSwap, while some appear to be similar to every other method. 

To summarize our insights from this experiment, the t-SNE results show that models with 2D and 3D convolutional learning capture different information. Spatial features learn strong method-specific attributes, while spatio-temporal features capture shared attributes between different but similar deepfake types.

\subsubsection{Relation of Features to Generalization}
The two distinct natures of clustering demonstrated by spatial and spatio-temporal features explain our findings in the generalization experiment. Since spatial features are very method-specific and do not associate even the learned methods with each other, they also fail to associate newly encountered deepfake types with already learned ones. However, the spatio-temporal features capture similarities between learned methods. Thus, when they encounter a new method, they can identify commonalities with the learned deepfakes and detect them more consistently than detectors that rely on spatial features; image and sequence encoders.

\subsection{Utilization of Temporal Information}\label{extime}
In this experiment, we investigate whether sequential and hierarchical video encoders, RNNs and 3D CNNs, are different in how they utilize the temporal information of deepfake videos. 

A typical evaluation technique for video encoders is to alter the video sequence in some ways, such as partially freezing it or reversing the order \citep{ManttariINTERPRETINGNETWORKS}. These experiments can provide
insights on how the encoders use the temporal information. Motivated by the video encoding experiments implemented for other tasks \citep{Ghodrati2018VideoEvaluation, ManttariINTERPRETINGNETWORKS}, we design an experiment suitable for our task.

\begin{figure}[t!]
\centering
\includegraphics[width=0.49\textwidth]{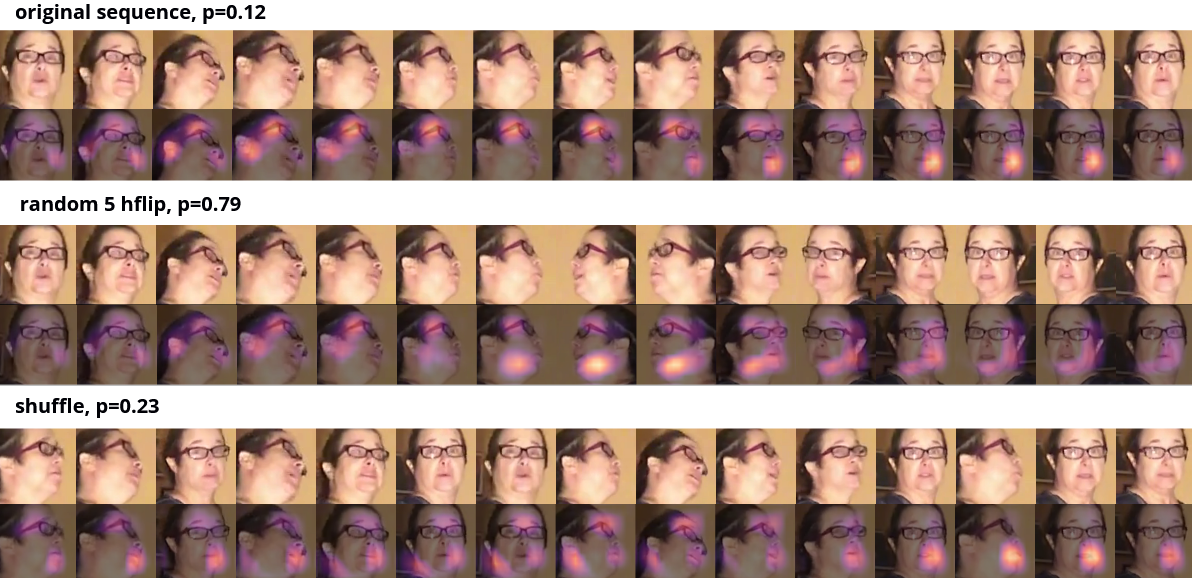}
\caption{Sequence alterations applied for R3D and Grad-CAM \citep{Selvaraju2016Grad-CAM:Localization} heatmaps. The true label is real. p is the model prediction; probability of being fake.} 
\label{fig:mirror}
\end{figure}

The way models use temporal information can be inferred from whether they depend on the sequence's temporal correlation, namely frame order, and alignment. We handpick a set of test samples from the DFDC dataset that have motion, pose change or completely side poses. We select these samples because they are not monotone like the average frontal and static portrait video; hence, the frames' order and alignment carry much more significance. DFDC is used for this experiment as FF++ has exclusively static and frontal videos. The models are trained on the DFDC train split and other datasets are not associated with this experiment. Here, we select the sequence encoder with an LSTM and R3D as the 3D video encoder, as these are the best performing models from each type of encoder on DFDC data. A sample from the handpicked set can be found in Figure \ref{fig:mirror} top row. To evaluate if the models require the frames to spatially correlate, we disrupt the alignment by horizontally 
flipping randomly selected frames. We start by flipping a single frame, gradually flip more frames, and finally flip every second frame to produce the most misaligned case.
Secondly, we test how crucial the order of the frames is by shuffling. Figure \ref{fig:mirror} shows random five frame horizontal flipping and shuffling. 

Results are provided in Table \ref{myexp}. We focus on the real-class loss (colored blue) because the misaligned and unordered data gets confused for fake-class. This is because the correlation of the frames is lost, leading to the recognition of false artifacts. Because of this, the fake metric can falsely improve. As we go over the results, we are not interested in how high or low loss values are; we are rather interested in how the model's loss changes when we apply the alterations.

The RNN is not noticeably affected by misaligned i.e. flipped samples. This is because as the RNN takes high-level features produced individually by the CNN and the misalignment between frames is not reflected in this abstraction. On top of that, the global sequence encoding loses the local spatial features. Hence, it cannot track the severe disruption in the correlation of spatial and temporal dimensions. Interestingly, RNN has also not been affected by shuffling. This shows that the RNN is indifferent to the temporal order in deepfake detection. The input spatial features indicate whether a timestep is real or fake-leaning. The rearrangement of the order of these signals does not change the overall frequency in any notable way. Depending on what is desired from RNNs, these attributes may seem advantageous and robust; however, in deepfake detection, we find it to be inadequate utilization of temporal information and correlation.

R3D exhibits higher loss as the sequence gets more misaligned and unordered. We see that it can tolerate minor errors like one or three flipped frames, as most of the clip
remains unaltered. This is because of the hierarchical processing of the temporal patches, which is characteristic of 3D convolutional architecture. Thus an error in a patch of frames does not reflect strongly onto the rest of the sequence. For the most misaligned case, R3D has log-loss much worse than random. R3D shows high loss also for shuffling but less than flipping every second frame. This is because the shuffled sequence can have frames that correspond by chance. 

In Figure \ref{fig:mirror}, we can see the effects of the alterations on R3D's predictions and activations. In the top row, we have the unaltered real sequence, and the Grad-CAM \citep{Selvaraju2016Grad-CAM:Localization} visualizations highlighting the fake suspicions. The suspicious regions correlate for consecutive temporal patches. When we apply some flipping, R3D activates around flipped frames depending on the severity of the misalignment. This pushes R3D to predict the sequence to be fake with higher probability, while it had been classified accurately as real before. 

To conclude the experiment, the results show that the 3D Convnet truly learns and relies on the spatio-temporal correlation of a video sequence. In contrast, the sequence encoder RNN is indifferent to the temporal order and local disruptions in spatio-temporal correlation in deepfake videos.

\begin{table*}[]\centering
\footnotesize
\caption{Experiment evaluating how changes in the frame sequence affect the model performance. Log-loss is shown on the table for a handpicked set of 95 real and 72 fake samples. Flipping is horizontal. Sequence length is 16. The RNN pipeline in use is LSTM on top of the EfficientNet-B3.}\label{myexp}
\begin{tabular}{l|c||c||c|c|c|c||c}
\hline
 &  &  & \multicolumn{4}{c||}{flipping n random frames} &  \\ \cline{4-7}
\multirow{-2}{*}{} & \multirow{-2}{*}{\begin{tabular}[c]{@{}c@{}}loss of  class\end{tabular}} & \multirow{-2}{*}{\begin{tabular}[c]{@{}c@{}}original\end{tabular}} & 1 & 3 & 5 & every 2nd & \multirow{-2}{*}{shuffle} \\ \hline
 & \cellcolor[HTML]{E7FFFE}real & \cellcolor[HTML]{E7FFFE}0.15 & \cellcolor[HTML]{E7FFFE}0.15 & \cellcolor[HTML]{E7FFFE}0.15 & \cellcolor[HTML]{E7FFFE}0.15 & \cellcolor[HTML]{E7FFFE}0.14 & \cellcolor[HTML]{E7FFFE}0.14 \\ \cline{2-8} 
\multirow{-2}{*}{RNN} & fake & 0.36 & 0.38 & 0.38 & 0.42 & 0.48 & 0.39 \\ \hline
 & \cellcolor[HTML]{E7FFFE}real & \cellcolor[HTML]{E7FFFE}0.32 & \cellcolor[HTML]{E7FFFE}0.33 & \cellcolor[HTML]{E7FFFE}0.41 & \cellcolor[HTML]{E7FFFE}0.58 & \cellcolor[HTML]{E7FFFE}1.66 & \cellcolor[HTML]{E7FFFE}0.65 \\ \cline{2-8} 
\multirow{-2}{*}{R3D} & fake & 0.48 & 0.46 & 0.35 & 0.25 & 0.16 & 0.25 \\ \hline
\end{tabular}
\end{table*}

\section{Conclusion}

In this work, we employed \textit{spatio-temporal learning} by 3D Convnets to bring a robust temporal modeling solution to generalizable deepfake-video detection. We evaluated our approach against the existing image and sequence encoders;  CNNs and RNNs. These two encoders represent\textit{ spatial features} and \textit{sequential dependencies between spatial features}. We analyzed all three models through extensive experimentation, where we reached the following conclusions:

\begin{enumerate}[(a)]
    \item Spatio-temporal features achieve generalization to different deepfake methods, while models that depend on spatial features fail in comparison.
    \item Deepfake datasets should be diverse in terms of the deepfake algorithms and tasks used; otherwise, if a unique method is encountered, even the generalizing models fail.
    \item Spatio-temporal features capture similarities between different deepfake methods, while spatial features capture method specific attributes.
    \item Sequence encoders are indifferent to order and local spatio-temporal inconsistencies in a deepfake video sequence while 3D Convnets model spatio-temporal correlation thus capture disruptions of it.
\end{enumerate}

In conclusion, we have uncovered that image encoders are not reliable nor robust detectors to deploy against new deepfake methods. Sequence encoders also fall short in this regard, as they depend on an image encoder for feature extraction. We have demonstrated that 3D convolutional models are the most effective and favorable models in terms of generalization and that they owe this to their modeling of spatio-temporal features.






\bibliographystyle{model2-names}
\bibliography{refs}

\section*{Supplementary Material}\label{supp}
Please see \ref{a1} and \ref{a2} for supplementary material. This covers further detail on data processing, training; optimizers, regularization, learning-rate schedulers and data augmentations. We also provide the full results for the generalization experiment which provide further context and proof for how we compute the reported values.

\clearpage
\appendix


\section{Further Details on Experimental Setup} \label{a1}

\subsection{Data filtering steps}
In the main work we stated that we apply two filtering steps to the face bounding boxes we retrieve, to clear false detections and scale errors. First, we check if all the bounding boxes overlap with the neighboring boxes by checking if intersection over union is non-zero. After we discard those that fail, we discard size outliers. We retrieve a list of width lengths of the square boxes and compute each length's absolute distance to the median. The distance is then divided by the median. Finally, we pick a threshold such that the values above are outliers. Through observing the numbers for a small, diverse set of samples, we chose this value to be 10. If the value is lower, some videos where the bounding box size changes a lot due to movement are incorrectly detected as outliers. We prefer allowing a few bad samples in training rather than losing diversity in data samples. 

\subsection{Model properties}
Here, on Table \ref{models}, we report some of the architectural properties of our five models.

\begin{table*}
\centering
\caption{Summary of attributes of the architectures that are relevant for comparison.}\label{models}
\begin{tabular}{@{}lccccc@{}}
\toprule
\multicolumn{1}{c}{model name} & model type & \multicolumn{1}{l}{\# of params} & \# of conv layers & resolution & pretrained on \\ \midrule
XceptionNet \citep{Rossler2019FaceForensics++:Images} & CNN & 20.809 M & 36 & 299 & ImageNet \\
EfficientNet-B3 \citep{TanEfficientNet:Networks} & CNN & 10.698 M & 25 & 224 & ImageNet \\ \midrule
EfficientNet-B3 + LSTM & CNN + RNN & 30.245 M & 25 & 224 & \multirow{2}{*}{\begin{tabular}[c]{@{}c@{}}EfficientNet component\\  is pretrained on task\end{tabular}} \\
EfficientNet-B3 + biGRU & CNN + RNN & 26.441 M & 25 & 224 &  \\ \midrule
R3D-18 \citep{Tran2017ARecognition} & 3D CNN & 33.167 M & 18 & 224 & Kinetics-400 \\
I3D-RGB \citep{CarreiraQuoDataset} & 3D CNN & 12.288 M & 22 & 224 & ImageNet/Charades \\ \bottomrule
\end{tabular}
\end{table*}

\subsection{Sampling input clips} We provide a more detailed description of how we sample input clips at training and inference. For each video, we acquire three frames per second. This results in $T\times3$ frames for a video that is  $T$ seconds long. Most of the time, $T\times3$ is more than 16 frames, which is the length we train our video-encoders with. During training, we pick a random 16-frame subsequence from the available frames. This means that at each epoch, a different part of the video will be seen. At inference, we take all the available frames and retrieve subsequences of length 16. In FaceForensics++, we use 16-frame skips in between, meaning there is no overlap. In DFDC \citep{Dolhansky2020TheDataset}, because videos are shorter, we use two frame skips, thus get overlapping subsequences. Batches of subsequences are forwarded through the video encoders, and the predictions of each set are averaged. If we can only retrieve less than 16 frames from a video, we loop the sequence at the end until we complete the length. 

\subsection{Training}
 We train the models until the validation loss diverges, overfits or stabilizes. We save the model with the best validation loss to use in the experiments. 

\paragraph{Optimization} For all models, we use Adam optimizer\cite{KingmaADAM:OPTIMIZATION}. As for the learning rate, we use 1e-5 for R3D and 2e-6 for recurrent nets with pretrained CNN backbones. Higher learning rates result in oscillating training loss for the video encoders, while lower results in slow learning. This is likely because of our unfortunately small batch sizes; 2 for recurrent models and 4 for R3D. Due to memory constraints, larger batch sizes are not possible for the video encoders. For image classifiers, the learning rate is 1e-4, and the batch size is 8; oscillation is not a problem. 
\paragraph{Regularization} To mitigate overfitting, we use L2-regularization and Dropout in some networks. We use a weight-decay of 1e-5 for image classifiers and recurrent models. EfficientNet has a dropout layer ($p=0.3$) before the output layer. We don't use Dropout in Xception. In recurrent models, we also use additional Dropout layers, as described in the methodology. R3D also does not utilize Dropout; we use a small weight-decay of 1e-7. During training, R3D does not exhibit overfitting until many later epochs. 

\paragraph{Schedulers} We use learning rate schedulers to reduce and stabilize the training at later epochs. We experimented with both plateau and multiplicative learning rate reduction. The former scheduler reduces the learning rate if the given metric stopped improving for the last $N$ validations. $N$ is referred to as patience, and we set it to five epochs. The multiplicative policy reduces the learning rate at epochs specified before-hand. For the epoch milestones, we have decided on 10. For both schedulers, we use 0.1 for the reducing factor. For our models, the scheduler choice doesn't change much. This is because the heuristically-selected multiplicative epochs somewhat match the plateau choices. However, either one should be applied.

\paragraph{Parameter Tuning}
As all our nets are pretrained, the networks did not require a high effort parameter tuning process. We found the proper learning rate and scheduler parameters by grid search. Different seed runs also do not show much variance because of the nature of our data loading; set and class balance, and also because of the networks being pretrained.

\paragraph{Augmentations}
We apply certain image augmentations in training to make models robust against different color and illumination conditions, compression and noise levels. During training, we flip every sample horizontally with 0.5 probability. We also pick a random additional augmentation, which we also apply with 0.5 probability. This list of augmentations includes random or corner-crops, JpegCompression, Gaussian noise, blur, down-scaling, random brightness, contrast, and color shifts. We implement the augmentations using the albumentations\footnote{https://albumentations.readthedocs.io/} library. The augmentations were selected considering the data distributions at hand and any adversarial filtering that could be applied to fool detectors that do not expect such input.

\section{Full Results of Generalization Experiment}\label{a2}
On Table \ref{general}, the full results for the generalization experiment runs can be found. For a single model, this shows first the original run with no methods left-out on top row, then the following 5 rows are runs where the blue-colored method is excluded. In the main work we only provided the colored boxes in this table, as them alone suffice for analysis. 
\begin{table*}\centering \small
\begin{tabular}{cl|ccccc|cc|c|c}
\hline
\multicolumn{1}{l}{} & \multicolumn{1}{c|}{} & DF & F2F & FS & NT & FShi & Real & \multicolumn{1}{l|}{Fake} & avg & drop \\ \hline
\multicolumn{1}{c|}{} &  & 99.29 & 97.86 & 97.86 & 95.71 & 95.71 & 100.00 & 97.29 & 98.64 & - \\ \cline{3-11} 
\multicolumn{1}{c|}{} &  & \cellcolor[HTML]{D9E9FF}{\color[HTML]{000000} 21.43} & 91.43 & 95.00 & 95.71 & 94.29 & 99.29 & 79.57 & 89.43 & -9.21 \\
\multicolumn{1}{c|}{} &  & 97.14 & \cellcolor[HTML]{D9E9FF}{\color[HTML]{000000} 12.14} & 98.57 & 93.57 & 97.14 & 99.29 & 79.71 & 89.50 & -9.14 \\
\multicolumn{1}{c|}{} &  & 98.57 & 99.29 & \cellcolor[HTML]{D9E9FF}{\color[HTML]{000000} 0.00} & 97.86 & 87.14 & 99.29 & 76.57 & 87.93 & -10.71 \\
\multicolumn{1}{c|}{} &  & 91.43 & 93.57 & 97.14 & \cellcolor[HTML]{D9E9FF}{\color[HTML]{000000} 0.00} & 88.57 & 100.00 & 74.14 & 87.07 & -11.57 \\
\multicolumn{1}{c|}{} &  & 95.71 & 88.57 & 99.29 & 93.57 & \cellcolor[HTML]{D9E9FF}1.43 & 100.00 & 75.71 & 87.86 & -10.78 \\ \cline{3-11} 
\multicolumn{1}{c|}{} & \multirow{-7}{*}{XceptionNet} & \multicolumn{8}{r|}{} & \cellcolor[HTML]{FFCCC9}-10.28 \\ \cline{2-11} 
\multicolumn{1}{c|}{} &  & 97.86 & 94.29 & 98.57 & 97.86 & 96.43 & 100.00 & 97.00 & 98.50 & - \\ \cline{3-11} 
\multicolumn{1}{c|}{} &  & \cellcolor[HTML]{D9E9FF}15.71 & 92.14 & 91.43 & 95.00 & 97.14 & 100.00 & 78.29 & 89.14 & -9.36 \\
\multicolumn{1}{c|}{} &  & 95.71 & \cellcolor[HTML]{D9E9FF}7.14 & 92.86 & 93.57 & 94.29 & 99.29 & 76.71 & 88.00 & -10.50 \\
\multicolumn{1}{c|}{} &  & 95.00 & 92.14 & \cellcolor[HTML]{D9E9FF}0.00 & 93.57 & 94.29 & 100.00 & 75.00 & 87.50 & -11.00 \\
\multicolumn{1}{c|}{} &  & 87.86 & 81.43 & 95.00 & \cellcolor[HTML]{D9E9FF}0.00 & 91.43 & 100.00 & 71.14 & 85.57 & -12.93 \\
\multicolumn{1}{c|}{} &  & 100.00 & 94.29 & 95.71 & 97.14 & \cellcolor[HTML]{D9E9FF}15.00 & 99.29 & 80.43 & 89.86 & -8.64 \\ \cline{3-11} 
\multicolumn{1}{c|}{\multirow{-14}{*}{2D}} & \multirow{-7}{*}{EfficientNet} & \multicolumn{8}{r|}{} & \cellcolor[HTML]{FFCCC9}-10.49 \\ \hline
\multicolumn{1}{c|}{} &  & 100.00 & 100.00 & 100.00 & 98.57 & 98.57 & 97.14 & 99.43 & 98.29 & - \\ \cline{3-11} 
\multicolumn{1}{c|}{} &  & \cellcolor[HTML]{D9E9FF}71.43 & 99.29 & 99.29 & 98.57 & 97.86 & 97.14 & 93.29 & 95.21 & -3.08 \\
\multicolumn{1}{c|}{} &  & 98.57 & \cellcolor[HTML]{D9E9FF}35.00 & 96.43 & 99.29 & 98.57 & 95.00 & 85.57 & 90.29 & -8.00 \\
\multicolumn{1}{c|}{} &  & 100.00 & 100.00 & \cellcolor[HTML]{D9E9FF}4.29 & 99.29 & 98.57 & 98.57 & 80.43 & 89.50 & -8.79 \\
\multicolumn{1}{c|}{} &  & 97.86 & 97.14 & 100.00 & \cellcolor[HTML]{D9E9FF}7.86 & 98.57 & 100.00 & 80.29 & 90.14 & -8.15 \\
\multicolumn{1}{c|}{} &  & 100.00 & 99.29 & 100.00 & 98.57 & \cellcolor[HTML]{D9E9FF}26.43 & 97.86 & 84.86 & 91.36 & -6.93 \\ \cline{3-11} 
\multicolumn{1}{c|}{\multirow{-7}{*}{\begin{tabular}[c]{@{}c@{}}2D\\ +\\ Seq.\end{tabular}}} & \multirow{-7}{*}{Eff + biGRU} & \multicolumn{8}{r|}{} & \cellcolor[HTML]{FFCCC9}-6.99 \\ \hline
\multicolumn{1}{c|}{} &  & 97.86 & 97.86 & 99.29 & 99.29 & 95.71 & 93.57 & 98.00 & 95.79 & - \\ \cline{3-11} 
\multicolumn{1}{c|}{} &  & \cellcolor[HTML]{D9E9FF}{\color[HTML]{000000} 62.14} & 100.00 & 99.29 & 100.00 & 98.57 & 90.71 & 92.00 & 91.36 & -4.43 \\
\multicolumn{1}{c|}{} &  & 97.86 & \cellcolor[HTML]{D9E9FF}{\color[HTML]{000000} 75.00} & 97.14 & 98.57 & 97.14 & 94.29 & 93.14 & 93.71 & -2.08 \\
\multicolumn{1}{c|}{} &  & 97.86 & 100.00 & \cellcolor[HTML]{D9E9FF}{\color[HTML]{000000} 5.00} & 100.00 & 96.43 & 92.86 & 79.86 & 86.36 & -9.43 \\
\multicolumn{1}{c|}{} &  & 97.86 & 97.14 & 98.57 & \cellcolor[HTML]{D9E9FF}{\color[HTML]{000000} 57.14} & 96.43 & 93.57 & 89.43 & 91.50 & -4.29 \\
\multicolumn{1}{c|}{} &  & 99.29 & 100.00 & 100.00 & 98.57 & \cellcolor[HTML]{D9E9FF}{\color[HTML]{000000} 56.43} & 92.14 & 90.86 & 91.50 & -4.29 \\ \cline{3-11} 
\multicolumn{1}{c|}{} & \multirow{-7}{*}{R3D \citep{Tran2017ARecognition}} & \multicolumn{8}{r|}{} & \cellcolor[HTML]{FFCCC9}-4.90 \\ \cline{2-11} 
\multicolumn{1}{c|}{} &  & 97.86 & 98.57 & 94.29 & 95.00 & 96.43 & 80.71 & 96.43 & 88.57 & - \\ \cline{3-11} 
\multicolumn{1}{c|}{} &  & \cellcolor[HTML]{D9E9FF}77.86 & 97.86 & 96.43 & 97.86 & 95.71 & 70.00 & 93.14 & 81.57 & -7.00 \\
\multicolumn{1}{c|}{} &  & 100.00 & \cellcolor[HTML]{D9E9FF}87.86 & 96.43 & 95.71 & 97.14 & 80.71 & 95.43 & 88.07 & -0.50 \\
\multicolumn{1}{c|}{} &  & 99.29 & 97.86 & \cellcolor[HTML]{D9E9FF}12.86 & 97.86 & 95.71 & 83.57 & 80.71 & 82.14 & -6.43 \\
\multicolumn{1}{c|}{} &  & 97.14 & 97.86 & 95.00 & \cellcolor[HTML]{D9E9FF}71.43 & 96.43 & 86.43 & 91.57 & 89.00 & 0.43 \\
\multicolumn{1}{c|}{} &  & 97.86 & 98.57 & 95.00 & 97.14 & \cellcolor[HTML]{D9E9FF}63.57 & 80.00 & 90.43 & 85.21 & -3.36 \\ \cline{3-11} 
\multicolumn{1}{c|}{\multirow{-14}{*}{3D}} & \multirow{-7}{*}{I3D-RGB  \citep{CarreiraQuoDataset}} & \multicolumn{8}{r|}{} & \cellcolor[HTML]{FFCCC9}-3.37 \\ \hline
\end{tabular}
\caption{Test precision per real and fake classes when one method is left out of training and validation (colored blue) but included in testing. Final column shows the drop in average accuracy, the red number is the average drop. The footnote symbols indicate shared attributes between methods. (DF: Deepfakes, F2F: Face2Face, FS: FaceSwap, NT: NeuralTextures, FShi: FaceShifter).}\label{general}
\end{table*}


\end{document}